\pdfoutput=1

\documentclass[11pt]{article}

\usepackage[final]{acl}

\usepackage{times}
\usepackage{latexsym}

\usepackage[T1]{fontenc}

\usepackage[utf8]{inputenc}

\usepackage{microtype}

\usepackage{inconsolata}

\usepackage{graphicx}
\usepackage{enumitem}
\usepackage{float}

\usepackage{algorithm}
\usepackage{algpseudocode}
\usepackage{amsmath}
\usepackage{booktabs}
\usepackage{listings}
\title{IFEval-Audio: Benchmarking Instruction-Following Capability in Audio-based Large Language Models}
\author{
Yiming Gao\textsuperscript{$\star$}\thanks{Work done during internship at I$^2$R and mentored by Bin Wang when he was there.}, 
Bin Wang\textsuperscript{$\dag$}, 
Chengwei Wei\textsuperscript{$\diamondsuit$}, 
Shuo Sun\textsuperscript{$\diamondsuit$}, 
AiTi Aw\textsuperscript{$\diamondsuit$}
\\
\textsuperscript{$\star$}Nanyang Technological University (NTU), Singapore \\
\textsuperscript{$\dag$}MiroMind \\
\textsuperscript{$\diamondsuit$}Institute for Infocomm Research (I$^2$R), A*STAR, Singapore\\
\texttt{GAOY0053@e.ntu.edu.sg} \\
}

\begin{document}

\maketitle
\begin{abstract}

Large language models (LLMs) have demonstrated strong instruction-following capabilities in text-based tasks. However, this ability often deteriorates in multimodal models after alignment with non-text modalities such as images or audio. While several recent efforts have investigated instruction-following performance in text and vision-language models, instruction-following in audio-based LLMs remains largely unexplored. To bridge this gap, we introduce \textbf{IFEval-Audio}, a novel evaluation dataset designed to assess the ability to follow instructions in an audio LLM. IFEval-Audio contains 280 audio–instruction–answer triples across six diverse dimensions: \textbf{Content, Capitalization, Symbol, List Structure, Length, and Format}. Each example pairs an audio input with a text instruction, requiring the model to generate an output that follows a specified structure. We benchmark state-of-the-art audio LLMs on their ability to follow audio-involved instructions. The dataset is released publicly to support future research in this emerging area.\footnote{Our evaluation code and datasets will be open-sourced at \url{https://github.com/AudioLLMs/AudioBench}}
\end{abstract}

\maketitle

\section{Introduction}\label{sec:instroduction}
Evaluation of LLMs has become a cornerstone of NLP research, with significant efforts dedicated to assessing their capabilities across diverse tasks~\citep{ye2023survey}. Instruction-following, defined as a model’s ability to produce outputs in a specified format or structure as per the given instruction~\citep{zhou2023ifeval}, has seen notable progress in text and image modalities through benchmarks like IFEval~\citep{zhou2023ifeval} and llava-Bench~\citep{chen2023llava}, as well as multimodal frameworks such as LAMM~\citep{liu2023lamm} and MM-IFEngine~\citep{ding2025mm_ifengine}. However, audio-based instruction-following remains largely underexplored. Existing audio datasets, such as MMAU~\citep{sakshi2024mmau} and SIFT-50M~\citep{pandey2025sift50m}, focus on tasks like speech recognition or audio understanding, but they rarely evaluate how well audio models adhere to complex instructions~\citep{moloo2017critical}. Moreover, while benchmarks like MIA-Bench~\citep{qian2024mia_bench} assess multimodal instruction-following for images, the instruction-following evaluation framework for audio modality is missing, limiting the development of audio LLMs for applications.

To address this gap, we introduce \textbf{IFEval-Audio}, a novel dataset designed to evaluate audio-based LLMs’ instruction-following capabilities, building on the call for innovative benchmarks~\citep{kiela2021dynabench}. It comprises 280 audio-instruction-answer triples across six dimensions, including Content, Capitalization, Symbol, Length, List Structure, and Format Requirements. Each triple pairs an audio input with a text instruction, requiring the model to generate a text output that adheres to the instruction’s specified format, such as producing a list or a single-sentence response. Unlike datasets like IFEval and LLaVA-Bench, which primarily assess format adherence without evaluating the correctness of the content, IFEval-Audio also evaluates the semantic correctness of the outputs, offering a more comprehensive assessment of model's instruction-following ability. We release IFEval-Audio publicly to advance audio instruction-following research.

\section{Related Work}\label{sec:related}
Instruction-following, the ability of LLMs to generate outputs in a specified format, has become a key focus in NLP research~\citep{zhou2023ifeval}. In text modalities, datasets like IFEval~\citep{zhou2023ifeval} set rigorous standards by using verifiable prompts to ensure models produce outputs such as numbered lists or single-sentence responses that meet structural requirements. Similarly, in image modalities, LLaVA~\citep{chen2023llava} evaluates models’ ability to describe visual content in prescribed formats, reflecting the maturity of instruction-following evaluation in these domains. However, these datasets and related multimodal efforts primarily focus on format adherence, lacking a comprehensive assessment of content correctness.

In the audio domain, datasets such as MMAU~\citep{sakshi2024mmau} and SIFT-50M~\citep{pandey2025sift50m} support tasks like speech recognition and audio understanding. Yet, they do not evaluate how well audio models follow complex instructions requiring structured text outputs, a gap underscored by critiques of current audio learning frameworks~\citep{moloo2017critical}. Recent work like Salmonn~\citep{tang2024salmonn} improves instruction-following in audio-based LLMs, but lacks a dedicated benchmark for structured instruction adherence. Unlike prior instruction-following datasets that focus solely on format, our benchmark, IFEval-Audio, introduces a dual evaluation approach, combining rule-based scoring for format adherence with LLM-based assessment for semantic correctness across six dimensions. This absence of audio-specific benchmarks highlights the need for IFEval-Audio to advance audio-based LLM development.

\section{IFEval-Audio}\label{sec:dataset}
The IFEval-Audio dataset is designed to evaluate the instruction-following capabilities of audio-based LLMs, focusing on both format adherence and semantic correctness. It comprises 280 audio-instruction-answer triples distributed across six dimensions, testing a model’s ability to generate text outputs that meet specific structural and content requirements based on audio inputs.

\subsection{Evaluation Dimensions}
IFEval-Audio spans six dimensions, each targeting a unique aspect of instruction-following capabilities.
\begin{enumerate}[itemsep=1pt, parsep=0pt, topsep=2pt, partopsep=0pt, leftmargin=10pt]
    \item \textbf{Content Requirements}: Instructions require including, removing, or replacing specific keywords (e.g., ``Include the word `technology' in your answer'').
    \item \textbf{Capitalization Requirements}: Instructions specify capitalization rules, such as all uppercase, all lowercase, or capitalizing specific words (e.g., ``Capitalize the first word of each sentence'').
    \item \textbf{Symbol Requirements}: Instructions mandate adding symbols at the start or end, quoting the output, or removing symbols (e.g., ``Enclose the response in square brackets'').
    \item \textbf{List Structure Requirements}: Instructions dictate list formats, including Arabic numerals, Roman numerals, English letter numerals, or bullet points (e.g., ``List items using Roman numerals'').
    \item \textbf{Length Requirements}: Instructions set word count constraints, such as upper limits, lower limits, or ranges (e.g., ``Respond within 10 words'').
    \item \textbf{Format Requirements}: Instructions require specific output formats, such as JSON (e.g., ``Provide the response in JSON format'').
\end{enumerate}

\subsection{Audio Sources and Diversity}
IFEval-Audio leverages audio from diverse datasets to ensure varied modalities and contexts, following the framework of AudioBench~\citep{wang2024audiobench}:
\begin{itemize}[itemsep=1pt, parsep=0pt, topsep=2pt, partopsep=0pt, leftmargin=10pt]
    \item \textbf{Speech Samples} are sourced from \texttt{Spoken SQuAD}~\citep{li2018spoken} (CC BY-SA 4.0), \texttt{TED-LIUM 3}~\citep{hernandez2018tedlium3} (CC BY-NC-ND 3.0), and \texttt{AudioBench}~\citep{wang2024audiobench} (inherited source licenses), providing conversational and question-answering contexts.
    \item \textbf{Music and Environmental Sound Samples} are drawn from \texttt{Muchomusic}~\citep{weck2024muchomusic} (CC BY-SA 4.0) and \texttt{Wavcaps}~\citep{mei2024wavcaps} (academic use only).
\end{itemize}
Instructions are designed to test one dimension each, varying in complexity. Answer formats range from lists to structured data like JSON. Speech audio (20--30 seconds) covers all six dimensions, while music/environmental sound audio (10 seconds) covers four dimensions (Content, Capitalization, Symbol, Format Requirements). Audio clips are used as-is from the source datasets without additional preprocessing for reproducibility.

\subsection{Dataset Structure}
The IFEval-Audio dataset is structured around 280 audio-instruction-answer triples, each comprising an audio input derived from the specified datasets, a text instruction targeting one of the six dimensions, and an expected answer that adheres to the instruction’s format and content requirements. The dataset distribution includes 240 triples from speech audio, with 40 triples per dimension across the six categories, and 40 triples from music and environmental sound audio, with 5 triples each per dimension for Content, Capitalization, Symbol, and Format Requirements (totaling 20 Music and 20 Environmental Sound). This organization ensures comprehensive coverage of instruction-following challenges across diverse audio modalities.

\begin{figure}[h]
    \centering
    \includegraphics[width=0.48\textwidth]{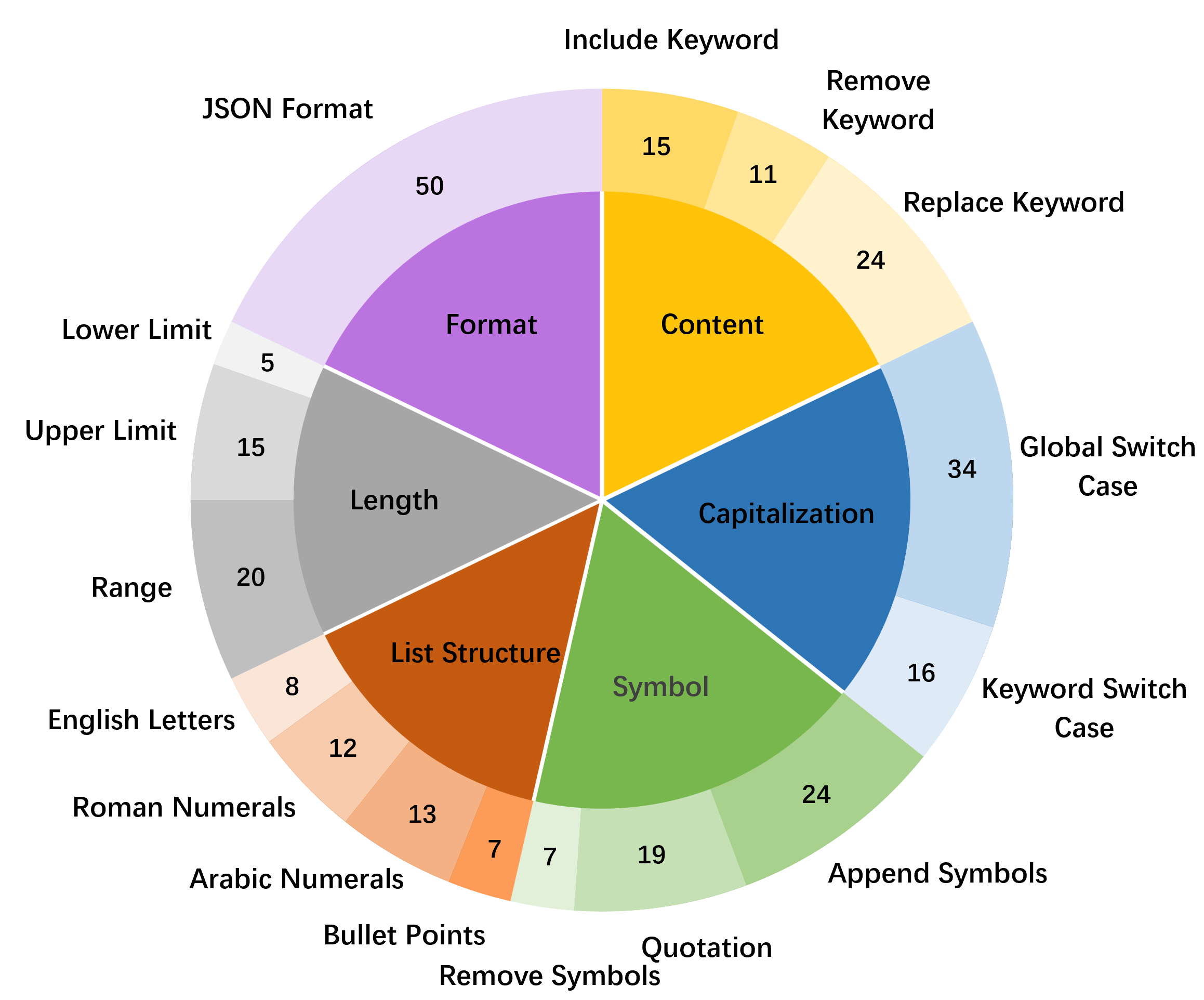}
    \caption{Breakdown of instruction subcategories within the dimensions, with approximately 25\% Content, 20\% Capitalization, 15\% Symbol, 15\% List Structure, 10\% Length, and 15\% Format requirements.}
    \label{fig:dimension_subcategories}
\end{figure}

Figure~\ref{fig:dimension_subcategories} shows the diversity of instruction types within the dimensions.

\subsection{Annotation Process}
The IFEval-Audio dataset was constructed through a meticulous process of curating audio clips from the specified sources, followed by manually designing instructions and answers to target each dimension comprehensively. This process involved selecting audio clips to represent diverse contexts, crafting instructions to test each of the six dimensions thoroughly, and creating expected answers that align with both the instructions and the audio inputs. By ensuring careful curation and design, this approach guarantees that IFEval-Audio serves as a robust benchmark for evaluating audio-based LLM instruction-following capabilities.

\subsection{Evaluation Design}
All triples in IFEval-Audio are designed to enable both rule-based and LLM-based evaluation for assessing instruction-following. A judge model evaluates each triple using the instruction, model output, and reference answer. Rule-based scoring applies strict rules to check format adherence (e.g., verifying a Roman numeral list), yielding a binary score (0/1) and a reason, which contributes to the Instruction Following Rate (IFR), the proportion of outputs correctly adhering to the specified format. LLM-based evaluation, using Meta Llama 3 70B~\citep{llama3modelcard}, assesses semantic correctness with a custom prompt template (detailed in Appendix~\ref{app:examples}). This produces a binary score (0/1) and a reason, which informs the Semantic Correctness Rate (SCR), the proportion of outputs semantically aligned with the reference answer. The Overall Success Rate (OSR) measures the proportion of outputs where both scores are 1 for a triple to contribute. Appendix~\ref{app:examples} provides a sample evaluation illustrating this process.

\section{Experiments}\label{sec:experiments}
This section presents the evaluation of audio-based LLMs on IFEval-Audio, focusing on results, patterns, findings, and implications.

\subsection{Models Tested}
Six models were evaluated, with the following aliases used in Table~\ref{tab:results}:
\begin{itemize}[itemsep=1pt, parsep=0pt, topsep=2pt, partopsep=0pt, leftmargin=10pt]
    \item \texttt{cascade-whisper-large-v3-llama-3-8b-Ins} (cascade): A custom model combining Whisper~\citep{pmlr-v202-radford23a} for speech recognition and Llama 3 8B Instruct~\citep{llama3modelcard} for language generation, with additional fine-tuning.
    \item \texttt{gemini-1.5-flash} (gemini-1.5): Google’s Gemini 1.5~\citep{gemini15_2024}, optimized for speed via the Google API.
    \item \texttt{gemini-2.0-flash-exp} (gemini-2.0): An experimental variant of Google’s Gemini models~\citep{gemini20_2024}, accessed via the Google API.
    \item \texttt{gpt-4o-audio-preview} (gpt4-audio): OpenAI’s GPT-4o~\citep{openai2024gpt4o_audio_docs} with audio capabilities, accessed via the OpenAI API.
    \item \texttt{Phi-4-multimodal-instruct} (phi-4): A multimodal instruction-following model based on Microsoft’s Phi series~\citep{phi4_2024}.
    \item \texttt{Qwen2-Audio-7B-Instruct} (qwen2): Alibaba’s Qwen2-7B~\citep{chu2024qwen2}, fine-tuned for audio-based instruction following.
\end{itemize}
Each was assessed using the framework in Section~\ref{sec:dataset}.

\subsection{Results, Analysis and Discussions}
\begin{table*}[h]
    \centering
    \resizebox{0.92\textwidth}{!}{
    \begin{tabular}{lc|cccccc}
        \toprule
        Dimension & Metric & cascade & gemini-1.5 & gemini-2.0 & gpt4-audio & phi-4 & qwen2 \\
        \midrule
                & SCR & 0.56 & \textbf{0.70} & 0.62 & \textbf{0.70} & 0.44 & 0.60 \\
        Content & IFR & 0.58 & \textbf{0.64} & 0.60 & 0.60 & 0.44 & 0.52 \\
                & OSR & 0.42 & \textbf{0.54} & 0.46 & 0.52 & 0.24 & 0.36 \\
        \midrule
                & SCR & 0.48 & \textbf{0.64} & 0.60 & 0.62 & 0.54 & 0.56 \\
        Capitalization & IFR      & 0.56 & \textbf{0.80} & 0.70 & 0.66 & 0.44 & 0.24 \\
                & OSR & 0.32 & \textbf{0.64} & 0.52 & 0.56 & 0.26 & 0.14 \\
        \midrule
                & SCR & 0.56 & \textbf{0.62} & 0.58 & 0.52 & 0.56 & 0.56 \\
        Symbol & IFR & 0.56 & 0.28 & \textbf{0.58} & 0.52 & 0.28 & 0.16 \\
                & OSR & 0.40 & 0.24 & \textbf{0.42} & 0.38 & 0.20 & 0.10 \\
        \midrule
                & SCR & 0.73 & 0.70 & 0.63 & \textbf{0.78} & 0.63 & 0.53 \\
        List Structure & IFR & 0.78 & 0.78 & 0.83 & \textbf{0.93} & 0.90 & 0.50 \\
                & OSR & 0.60 & 0.55 & 0.55 & \textbf{0.75} & 0.58 & 0.30 \\
        \midrule
                & SCR & 0.45 & \textbf{0.65} & 0.55 & 0.60 & 0.43 & 0.43 \\
        Length & IFR & 0.28 & \textbf{0.50} & 0.38 & 0.45 & 0.28 & 0.20 \\
                & OSR & 0.25 & \textbf{0.50} & 0.38 & 0.45 & 0.25 & 0.18 \\
        \midrule
                & SCR & 0.52 & 0.58 & 0.64 & \textbf{0.70} & 0.60 & 0.46 \\
        Format & IFR & 0.76 & 0.92 & \textbf{0.94} & 0.92 & 0.88 & 0.22 \\
                & OSR & 0.48 & 0.52 & 0.58 & \textbf{0.66} & 0.56 & 0.08 \\
                \midrule
        \multicolumn{2}{c |}{ \textbf{Overall Instruction-Following Rate}} & 0.59 & 0.65 & 0.67 & \textbf{0.68} & 0.53 & 0.30     \\
        \bottomrule
    \end{tabular}
    }
    \caption{Detailed performance across dimensions: Semantic Correctness Rate (SCR), Instruction Following Rate (IFR), and Overall Success Rate (OSR); Overall Instruction-Following Rate included at bottom.}
    \label{tab:results}
\end{table*}

Table~\ref{tab:results} highlights instruction-following capabilities across IFEval-Audio’s six dimensions. In \textbf{Content}, gemini-1.5 leads with an IFR of 0.64, reflecting its ability to incorporate keywords as instructed, likely due to extensive multimodal training inferred from Google’s data integration efforts. Conversely, phi-4’s low IFR (0.44) suggests struggles in following content instructions, possibly due to limited audio training, as its documentation emphasizes text efficiency. \textbf{Capitalization} shows gemini-1.5’s high IFR (0.80), indicating strong adherence to formatting rules, a strength in proprietary models, while qwen2’s IFR (0.24) reflects challenges despite its enhanced instruction-following design, possibly due to its focus on natural language prompts over specific formatting tasks. \textbf{Symbol} reveals varied instruction-following, with gemini-2.0’s IFR (0.58) suggesting experimental optimizations, but qwen2’s low IFR (0.16) indicates limited symbol training. \textbf{List Structure} favors gpt4-audio (IFR 0.93), excelling in structured instruction adherence, likely from structured text training, while qwen2’s IFR (0.50) shows moderate performance. \textbf{Length} challenges instruction following, with gemini-1.5’s IFR (0.50) reflecting training on length-constrained tasks, while qwen2’s IFR (0.20) suggests difficulties in constraint handling. \textbf{Format} sees gpt4-audio’s high IFR (0.92), likely from code training enabling structured output adherence, while qwen2’s IFR (0.22) highlights struggles with complex formats, as it outputs did not use double quotes " as standard JSON format.

IFEval-Audio underscores instruction-following gaps, particularly in Capitalization (IFR 0.24–0.80) and Format (IFR 0.22–0.94), urging improved multimodal alignment. gpt4-audio (overall instruction-following rate 0.68) and gemini-2.0 (0.67) excel with proprietary resources, while qwen2 (0.30) and phi-4 (0.53) indicate open-sourced models need targeted enhancements. As a benchmark, IFEval-Audio promotes research into fine-tuning and audio-text integration. For instance, in Format, gpt4-audio’s IFR (0.92) far exceeds its SCR (0.70), while qwen2’s IFR (0.22) aligns closer to its SCR (0.46), illustrating a mismatch between instruction-following and understanding capabilities across models.

\section{Conclusion}\label{sec:conclusion}
This study introduces IFEval-Audio, a novel dataset comprising 280 audio-instruction-answer triples across six dimensions, designed to evaluate the instruction-following capabilities of audio-based LLMs. By assessing both format adherence and semantic correctness, IFEval-Audio addresses a critical gap in multimodal NLP research, where audio instruction-following has been underexplored compared to text and image modalities. Experimental results on six state-of-the-art models highlight performance variations, with gpt4-audio and gemini-1.5 leading, while exposing weaknesses in handling complex formats. The public release of IFEval-Audio aims to foster further innovation, encouraging the development of robust audio-language models. 

\section{Limitations}\label{sec:limitations}
While IFEval-Audio provides a robust benchmark for audio instruction-following, it has notable limitations. The dataset contains 280 human-crafted triples, which may not fully capture the complexity of real-world audio scenarios. Additionally, the audio is English-only, limiting its applicability to multilingual contexts. The audio variety, though diverse across speech, music, and environmental sounds, is constrained by the sourced datasets (\texttt{Spoken SQuAD}, \texttt{TED-LIUM 3}, etc.), potentially missing broader acoustic diversity such as non-English speech or rare environmental sounds.

Evaluation methods also present constraints. The LLM-based semantic evaluation, relying on models like Meta Llama 3 70B~\citep{llama3modelcard}, introduces dependency on the judge model’s biases and limitations in understanding audio-derived context. Furthermore, the evaluation does not explicitly disentangle ASR or perception errors from instruction-following failures in SCR, potentially allowing transcription inaccuracies to dominate semantic assessments. In the future, we plan to expand the dataset’s scale, linguistic diversity, and improve evaluation robustness to better reflect real-world audio instruction-following challenges.

\bibliography{custom}

\appendix

\section{Additional Related Work}
Broader efforts in LLM evaluation, such as those surveyed by Ye et al.~\citep{ye2023survey}, emphasize the need for innovative benchmarks to address emerging challenges in NLP. Similarly, Dynabench~\citep{kiela2021dynabench} advocates for dynamic benchmarking to keep pace with evolving model capabilities. Other multimodal frameworks, including LAMM~\citep{liu2023lamm} and MM-IFEngine~\citep{ding2025mm_ifengine}, extend instruction-following to diverse modalities, yet their primary focus remains on format adherence rather than content correctness. Multimodal benchmarks, such as MIA-Bench~\citep{qian2024mia_bench}, have begun to explore instruction-following across modalities, but their inclusion of audio remains limited. These works underscore the importance of developing specialized datasets like IFEval-Audio to fill gaps in underrepresented domains like audio instruction-following.

In summary, while instruction-following has been extensively studied in text and image modalities, and audio datasets support related tasks, there is a clear gap in evaluating audio models’ ability to follow instructions with respect to both format adherence and semantic correctness. The detailed contributions of IFEval-Audio, as a novel dataset tailored to benchmark audio-based LLMs, offer a robust evaluation framework for advancing multimodal NLP research.

\section{Appendix Figures}
\begin{figure}[h]
    \centering
    \includegraphics[width=0.48\textwidth]{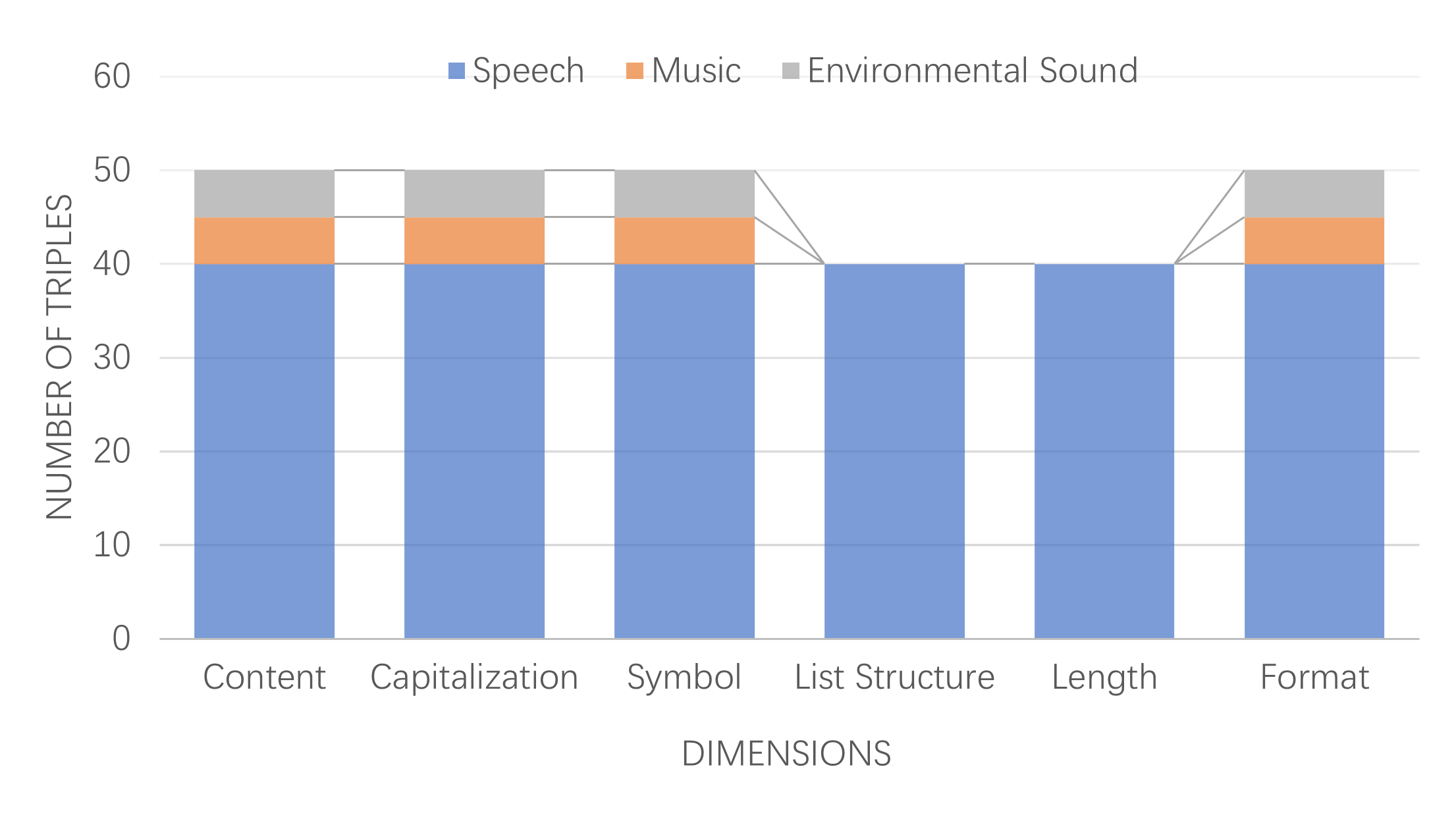}
    \caption{Distribution of triples across dimensions for Speech, Music, and Environmental Sound audio.}
    \label{fig:triples_per_dimension}
\end{figure}

Figure~\ref{fig:triples_per_dimension} illustrates the distribution of triples across dimensions.

\section{Appendix Examples}\label{app:examples}
Below is a sample evaluation of \textbf{cascade-whisper-large-v3-llama-3-8b-Ins} from the IFEval-Audio dataset, illustrating the evaluation process for a Content Requirements dimension triple.

\subsection{Sample Triple}
The original audio input is sourced from a sports commentary clip: ``with Rivera having been a linebacker with the Chicago Bears in Super Bowl XX, and Kubiak replacing Elway at the end of the Broncos' defeats in Super Bowls XXI and XXIV. This will be the first Super Bowl in which both head coaches played in the game themselves.'' Each triple in IFEval-Audio consists of an audio input, a text instruction, and an expected answer. The following triple targets the Content Requirements dimension:

\begin{lstlisting}
{
    "text": "What role did Rivera play in Super Bowl XX, and which team did he represent? Replace all occurrences of 'Super Bowl' with 'Championship Game' in your output.",
    
    "answer": "Rivera played as a linebacker for the Chicago Bears in Championship Game XX. His role on the team contributed to the Bears' participation in that game, marking his experience as a player in the Championship Game.",
    
    "dimension": "Content Requirements",
    
    "rule_type": "Replace Keyword",
    
    "rule_target": "Super Bowl",
    
    "task_type": "Content Requirements",
    
    "model_prediction": "Ron Rivera played as a linebacker for the Chicago Bears in Championship Game XX."
}
\end{lstlisting}

\subsection{Judge LLM Prompt Template}
The semantic correctness evaluation is performed using the Meta Llama 3 70B model as a judge. The following prompt template is used for each triple, with placeholders filled according to the triple evaluated.

\begin{lstlisting}
[Reference Answer]
{reference}
[Model Answer]
{prediction}
[Question]
{question}
[Task]
Rate the model's answer based on its alignment with the reference answer, focusing on the following two aspects:
1. **Correctness**: Assess if the model's answer demonstrates the correct understanding and response based on the [Reference Answer].
    Score 0: If the question is regarding to transcriptions, the model's answer is not the same as [Reference Answer]. If the question is not regarding to transcriptions, the model's answer does not accurately reflect the meaning or idea of [Reference Answer].
    Score 1: If the question is regarding to transcriptions, the model's answer is exactly the same as [Reference Answer]. If the question is not regarding to transcriptions, the model's answer accurately reflects the meaning or idea of [Reference Answer].
   
Please provide two separate ratings:
1. **Correctness Rating**: (0 or 1)
Your final output should be exactly in this format, you can only modify contents inside brackets.
Correctness Rating: (int)
Explanation: (Provide a concise explanation for each rating. For **Correctness**, explain if the model's answer is correct and aligns with the reference. For **Instruction-Following**, explain how well the model adhered to the task instructions and any discrepancies.)
\end{lstlisting}

Note that the Instruction-Following Rating is computed rule-based via a separate function (\texttt{format\_test}), yielding 1 if the prediction adheres to the specified rule (e.g., format, symbols), else 0. The overall success is 1 only if both ratings are 1. The decoding parameters for the judge include max\_tokens=512, n=1, with deterministic settings. The complete evaluation code, including this prompt and the \texttt{llama3\_70b\_as\_judge\_binary} function, is released on GitHub.

\subsection{Evaluation of the Triple}
The evaluation process involves both rule-based scoring for format adherence and LLM-based scoring for semantic correctness. The following is the evaluation of the prediction of the model against the reference answer:

\begin{lstlisting}
{
    "question": "What role did Rivera play in Super Bowl XX, and which team did he represent? Replace all occurrences of 'Super Bowl' with 'Championship Game' in your output.",
    "reference": "Rivera played as a linebacker for the Chicago Bears in Championship Game XX. His role on the team contributed to the Bears' participation in that game, marking his experience as a player in the Championship Game.",
    
    "model_prediction": "Ron Rivera played as a linebacker for the Chicago Bears in Championship Game XX.",
    
    "judge_response": "Correctness Rating: 1\nExplanation: The model's answer accurately reflects the meaning of the Reference Answer, correctly stating Rivera's role as a linebacker and his representation of the Chicago Bears in Championship Game XX. The answer is concise and directly answers the question.",
    
    "correctness_rating": 1,
    
    "instruction_following_rating": 1,
    
    "success": 1,
    
    "dimension": "Content Requirements"
}
\end{lstlisting}

The format score is 1, as the model correctly replaced ``Super Bowl'' with ``Championship Game'' according to the instruction. The semantic score is also 1, as the model’s output aligns with the reference answer in meaning and factual accuracy, despite minor stylistic differences (e.g., inclusion of ``Ron'').

\end{document}